\documentclass[conference]{IEEEtran}
\IEEEoverridecommandlockouts

\usepackage{cite}
\usepackage{amsmath,amssymb,amsfonts}
\usepackage{algorithmic}
\usepackage{graphicx}
\usepackage{textcomp}
\usepackage{xcolor}
\usepackage{floatrow}
\usepackage{tabularx,booktabs}
\usepackage{tabularx}

\def\BibTeX{{\rm B\kern-.05em{\sc i\kern-.025em b}\kern-.08em
    T\kern-.1667em\lower.7ex\hbox{E}\kern-.125emX}}
\begin{document}

\title{Content-based Analysis of the Cultural Differences between TikTok and Douyin}

\author{\IEEEauthorblockN{Li Sun*}
\IEEEauthorblockA{\textit{Georgen Institute for Data Science} \\
\textit{University of Rochester}\\
Rochester, United States \\
lsun12@u.rochester.edu}

\\

\IEEEauthorblockN{Songyang Zhang}
\IEEEauthorblockA{\textit{Department of Computer Science} \\
\textit{University of Rochester}\\
Rochester, United States \\
szhang83@ur.rochester.edu}

\and

\IEEEauthorblockN{Haoqi Zhang*\thanks{*Both authors contributed equally on this paper.}}

\IEEEauthorblockA{\textit{Department of Computer Science} \\
\textit{University of Rochester}\\
Rochester, United States \\
hzhang84@u.rochester.edu}

\\

\IEEEauthorblockN{Jiebo Luo}
\IEEEauthorblockA{\textit{Department of Computer Science} \\
\textit{University of Rochester}\\
Rochester, United States \\
jluo@cs.rochester.edu}

}

\maketitle

\begin{abstract}
  Short-form video social media shifts away from the traditional media paradigm by telling the audience a dynamic story to attract their attention. In particular, different combinations of everyday objects can be employed to represent a unique scene that is both interesting and understandable. Offered by the same company, TikTok and Douyin are popular examples of such new media that has become popular in recent years, while being tailored for different markets (e.g. the United States and China). The hypothesis that they express cultural differences together with media fashion and social idiosyncrasy is the primary target of our research. To that end, we first employ the Faster Regional Convolutional Neural Network (Faster R-CNN) pre-trained with the Microsoft Common Objects in COntext (MS-COCO) dataset to perform object detection. Based on a suite of objects detected from videos, we perform statistical analysis including label statistics, label similarity, and label-person distribution. We further use the Two-Stream Inflated 3D ConvNet (I3D) pre-trained with the Kinetics dataset to categorize and analyze human actions. By comparing the distributional results of TikTok and Douyin, we uncover a wealth of similarity and contrast between the two closely related video social media platforms along the content dimensions of object quantity, object categories, and human action categories. 
\end{abstract}


\begin{figure*}[t]
	\centering
	\begin{minipage}{.47\columnwidth}
		\centering
		\includegraphics[width=\textwidth]{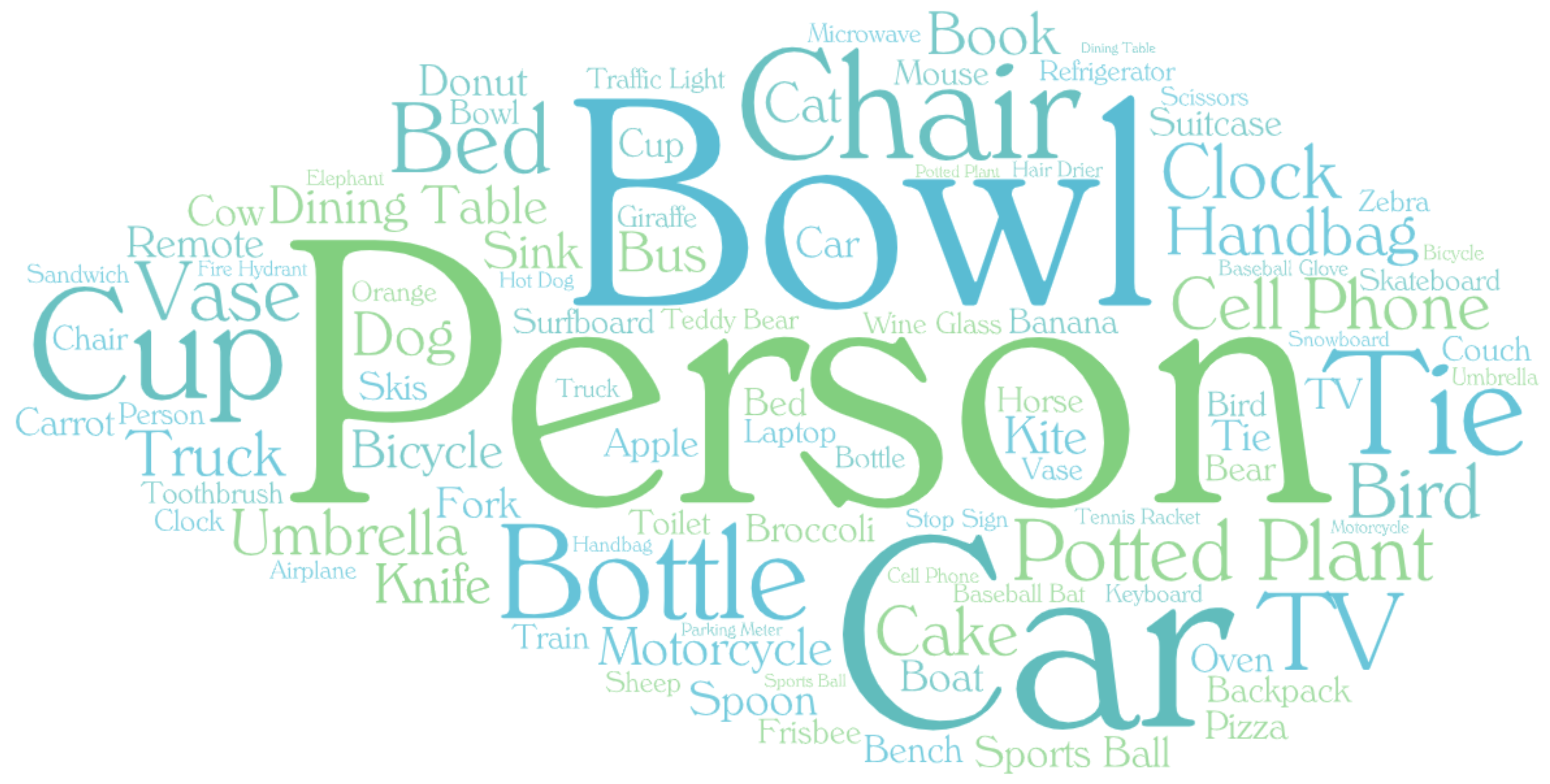} (a) Douyin
	\end{minipage}%
	\begin{minipage}{.47\columnwidth}
		\centering
		\includegraphics[width=\textwidth]{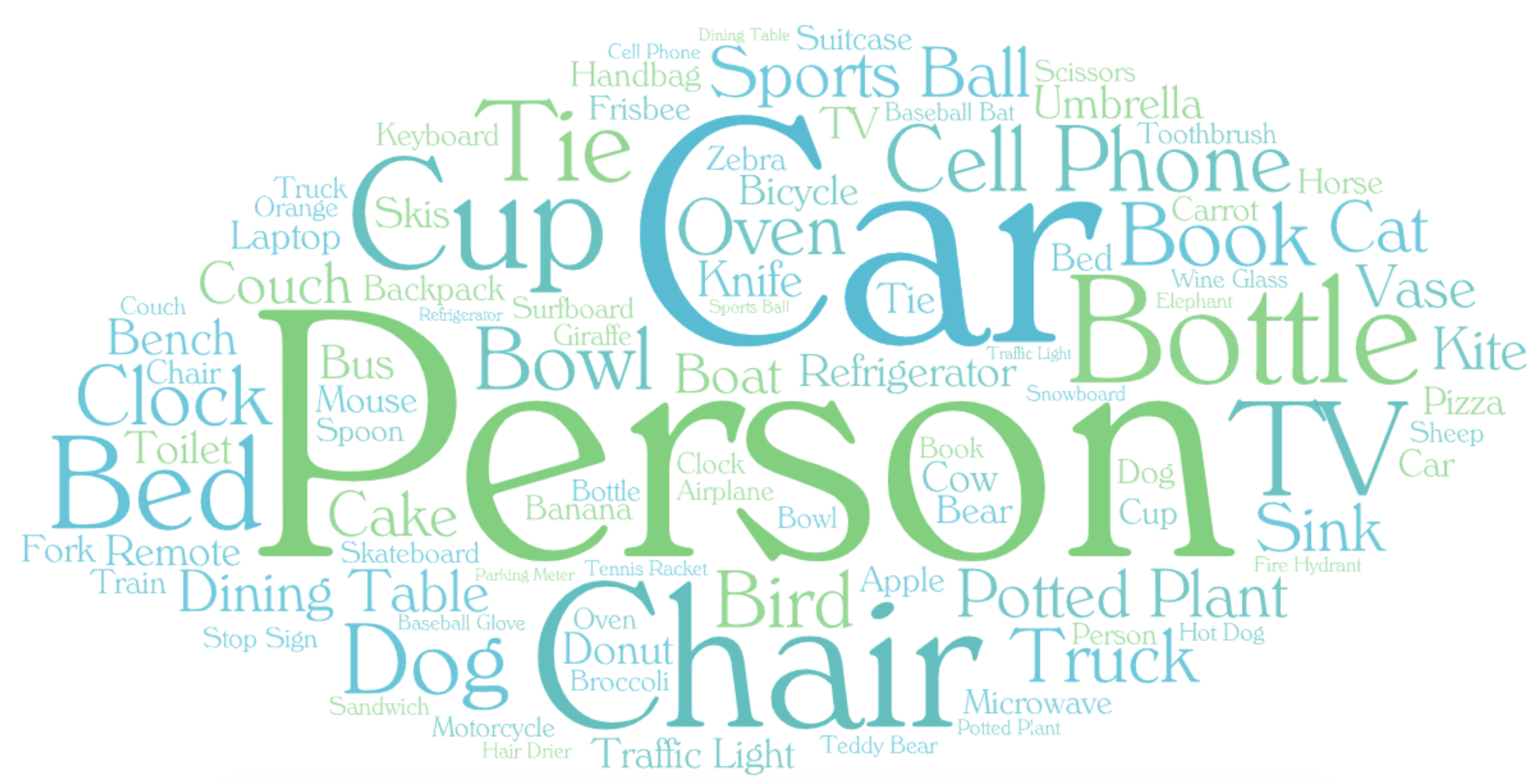} (b) TikTok
		\caption{Tag clouds of object labels.}
	\end{minipage}
\end{figure*}

\begin{figure*}[t]
	\centering
	\begin{minipage}{.47\columnwidth} 
		\centering
		\includegraphics[width=\textwidth]{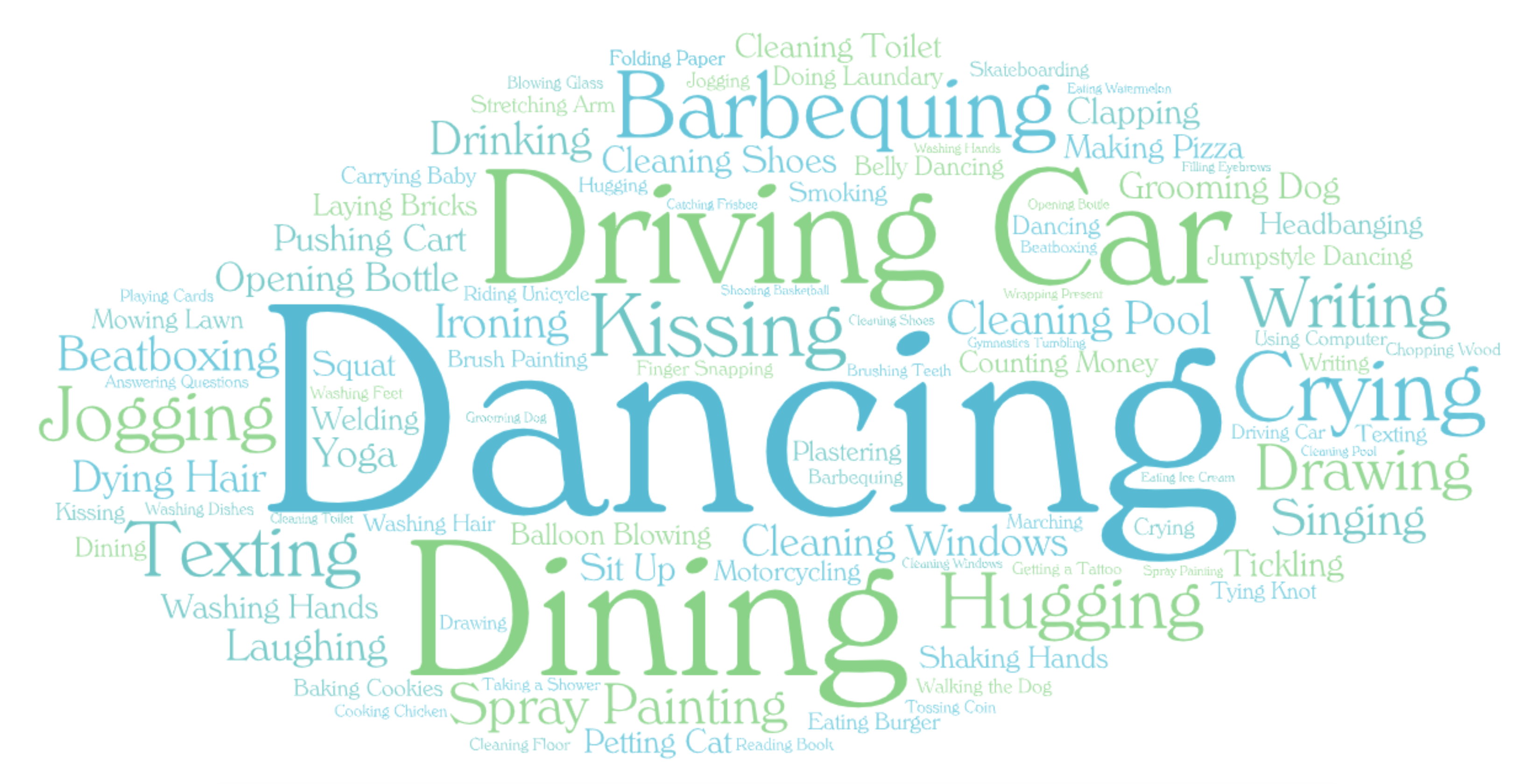} (a) Douyin
	\end{minipage}%
	\begin{minipage}{.47\columnwidth} 
		\centering
		\includegraphics[width=\textwidth]{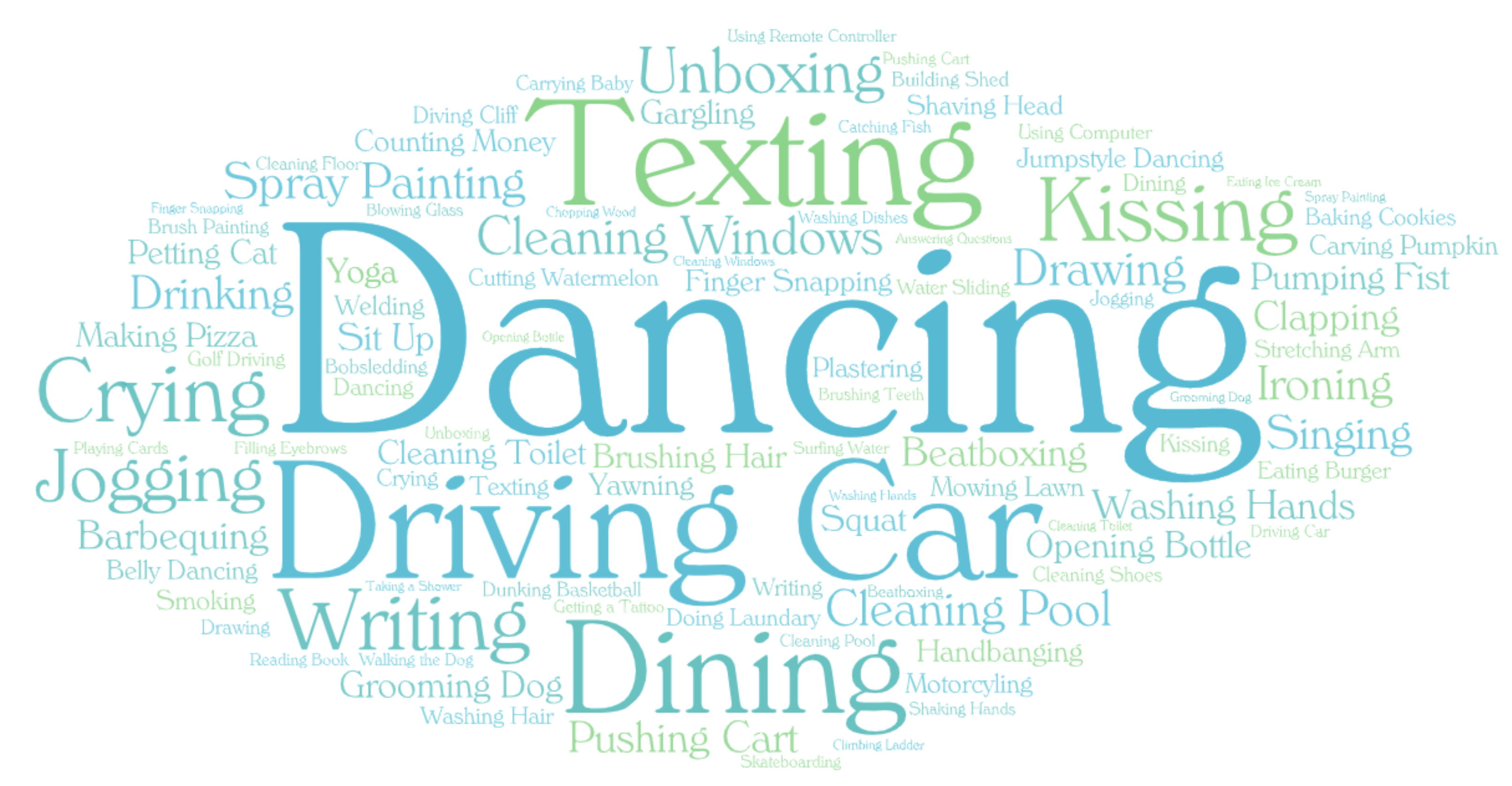} (b) TikTok
		\caption{Tag clouds of human actions.}
	\end{minipage}
\end{figure*}

\begin{table*}[t]
    \centering
    \caption{Taxonomy for Label Analysis.}
    \label{tab:the_table}
    \begin{tabular}{|c|c|c|}
        \hline
\textbf{Category}&{\textbf{Labels}}\\
\hline
Person & {\itshape Person}\\
\hline
Vehicle & {\itshape Bicycle, Car, Motorcycle, Airplane, Bus, Train, Truck, Boat}\\
\hline
Outdoor & {\itshape Traffic Light, Fire Hydrant, Stop Sign, Parking Meter, Bench}\\
\hline
Animal & {\itshape Bird, Cat, Dog, Horse, Sheep, Cow, Elephant, Bear, Zebra, Giraffe}\\
\hline
Accessory & {\itshape Backpack, Umbrella, Handbag, Tie, Suitcase}\\
\hline
\begin{tabular}{l}Sports\end{tabular} & {\itshape Frisbee, Skis, Snowboard, Sports Ball, Kite, Baseball Bat, Baseball Glove, Skateboard, Surfboard,}\\
& {\itshape Tennis Racket}\\
\hline
Tableware & {\itshape Bottle, Wine Glass, Cup, Fork, Knife, Spoon, Bowl}\\
\hline
Food & {\itshape Banana, Apple, Sandwich, Orange, Broccoli, Carrot, Hot Dog, Pizza, Donut, Cake}\\
\hline
Furniture & {\itshape Chair, Couch, Potted Plant, Bed, Dining Table, Toilet}\\
\hline
Electronic & {\itshape TV, Laptop, Mouse, Remote, Keyboard, Cell Phone, Microwave, Oven, Sink, Refrigerator}\\
\hline
Indoor & {\itshape Book, Clock, Vase, Scissors, Teddy Bear, Hair Drier, Toothbrush}\\
\hline
    \end{tabular}
\end{table*}

\begin{figure*}[t]
	\centering
	\begin{minipage}{\columnwidth}
		\centering
		\includegraphics[width=\textwidth]{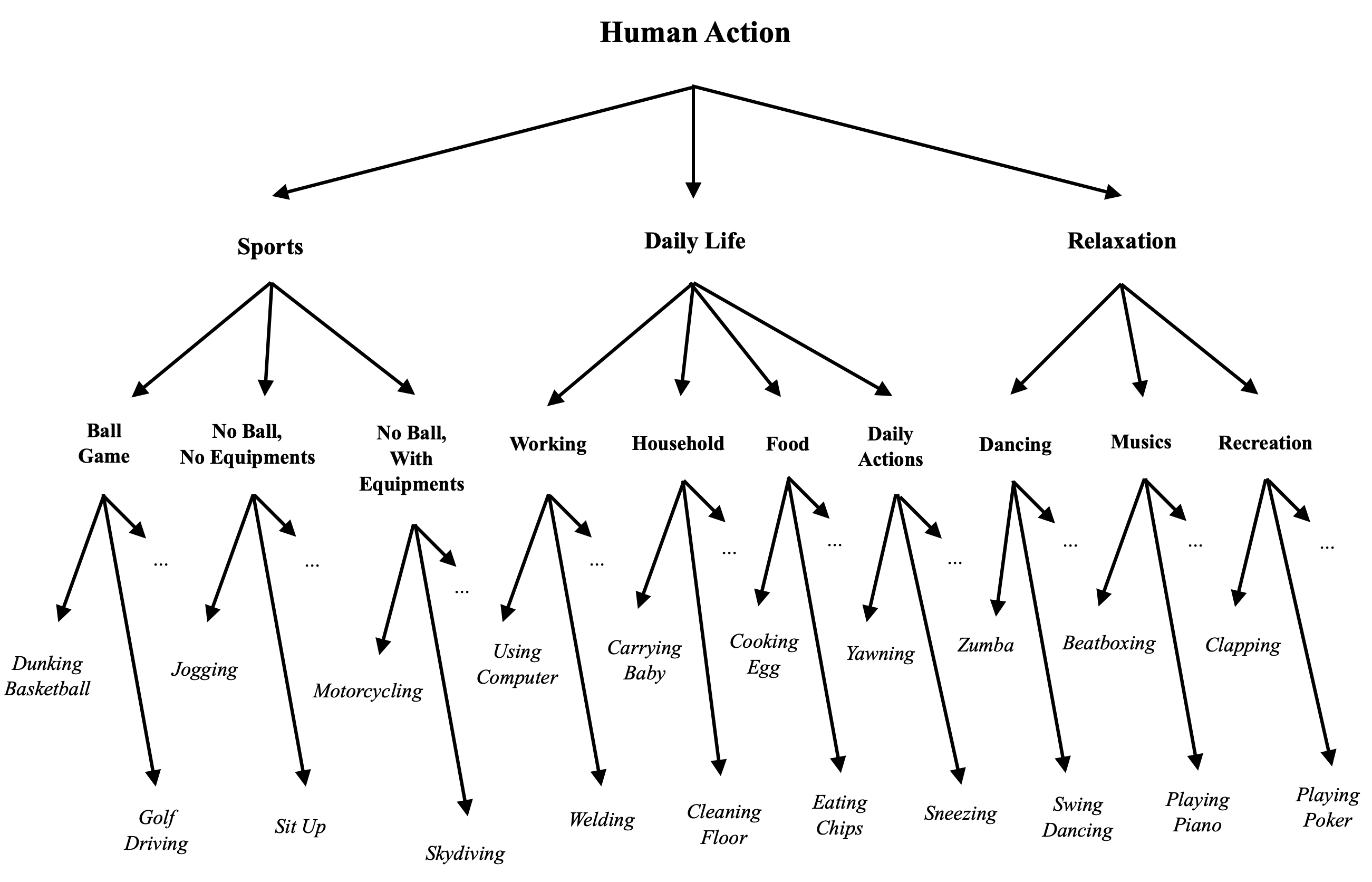}
		\caption{Taxonomy for human action analysis.}
	\end{minipage}
\end{figure*}

\begin{figure*}[t]
	\centering
	\begin{minipage}{.5\columnwidth}
		\centering
		\leftskip-2.7cm
		\includegraphics[height=17cm, keepaspectratio]{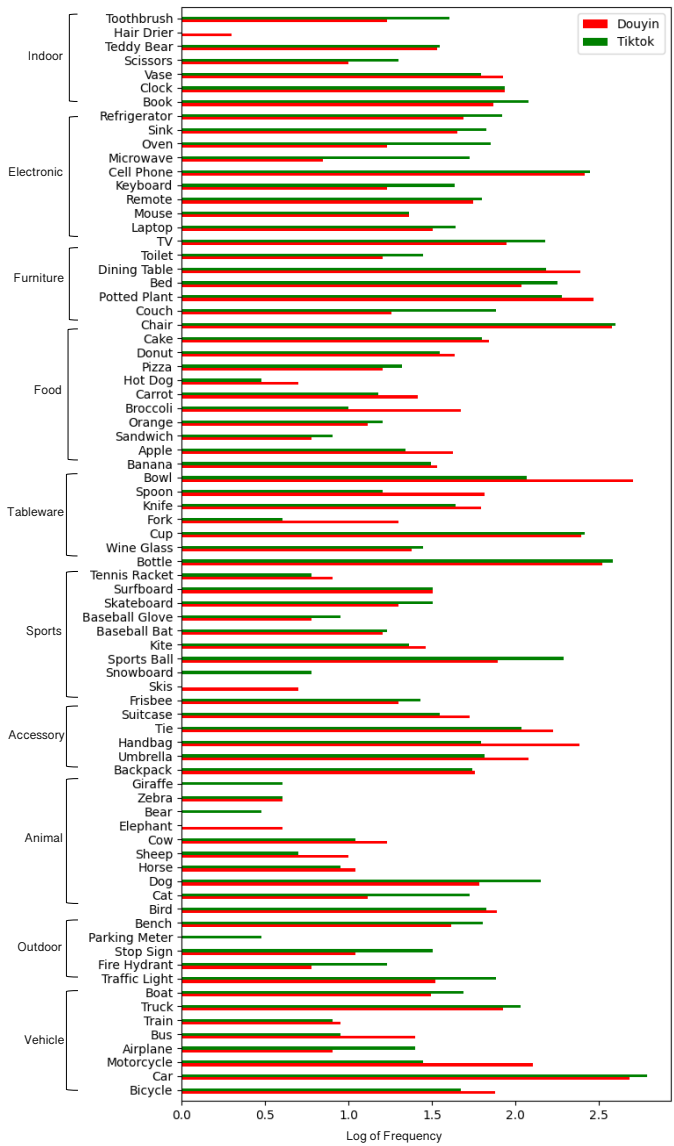}
	\end{minipage}%
	\begin{minipage}{.5\columnwidth}
		\centering
		\leftskip-2cm
		\includegraphics[height=17cm, keepaspectratio]{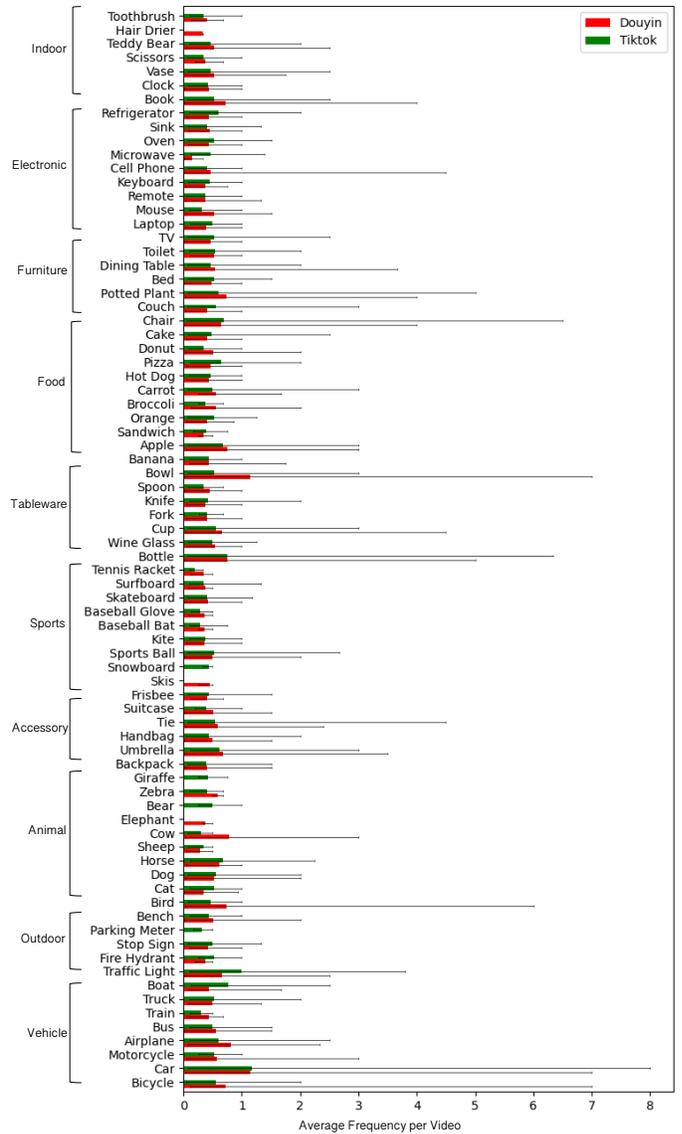}
	\end{minipage}
			\vspace{-0.4cm}
		\caption{Label frequency, mean, and range comparison for Douyin and TikTok (without {\itshape person}).}
\end{figure*}

\begin{figure*}[t]
	\centering
	\begin{minipage}{.49\columnwidth} 
		\centering
		\includegraphics[width=\textwidth, height=7cm, keepaspectratio]{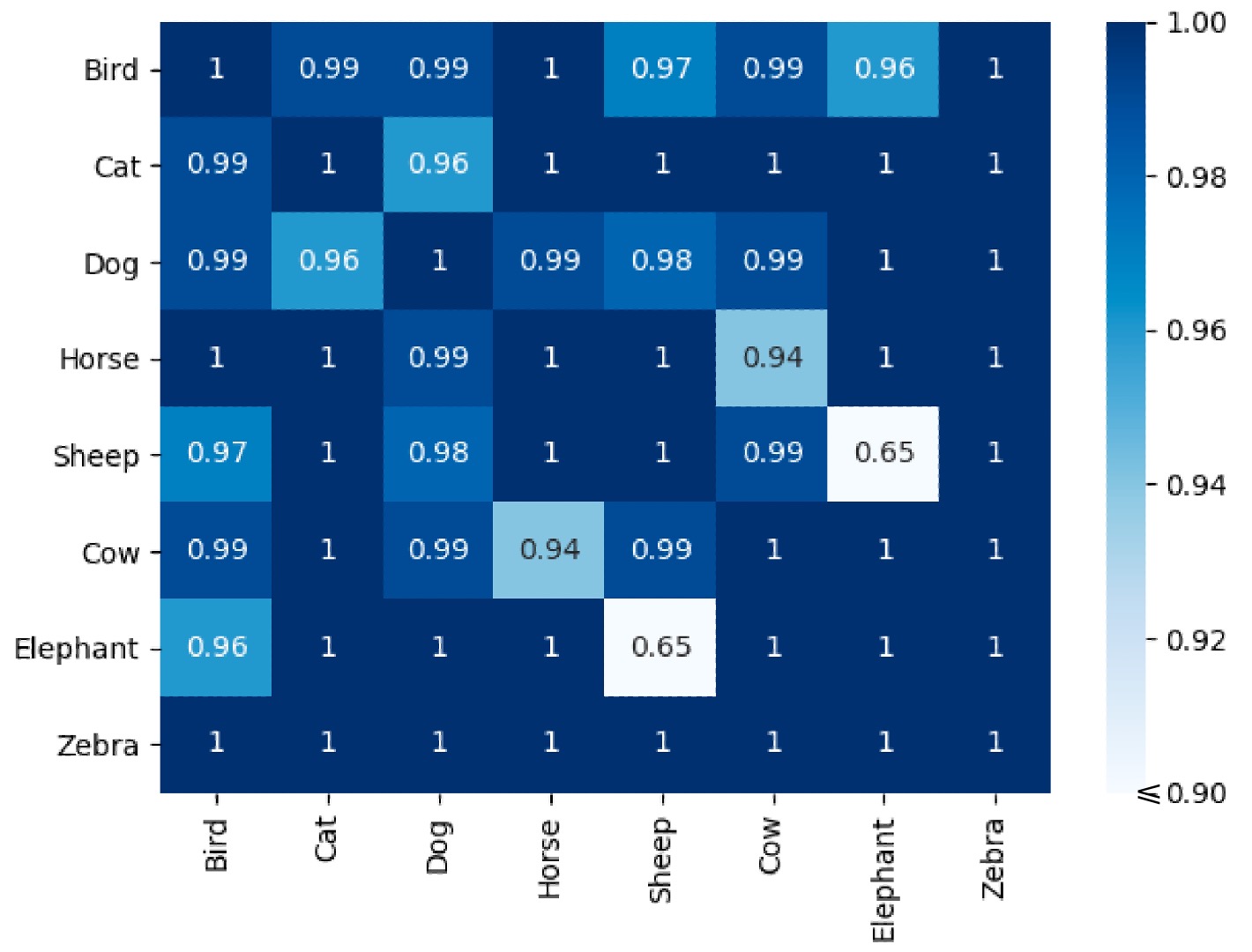} (a) Douyin
	\end{minipage}%
	\begin{minipage}{.49\columnwidth} 
		\centering
		\includegraphics[width=\textwidth, height=7cm, keepaspectratio]{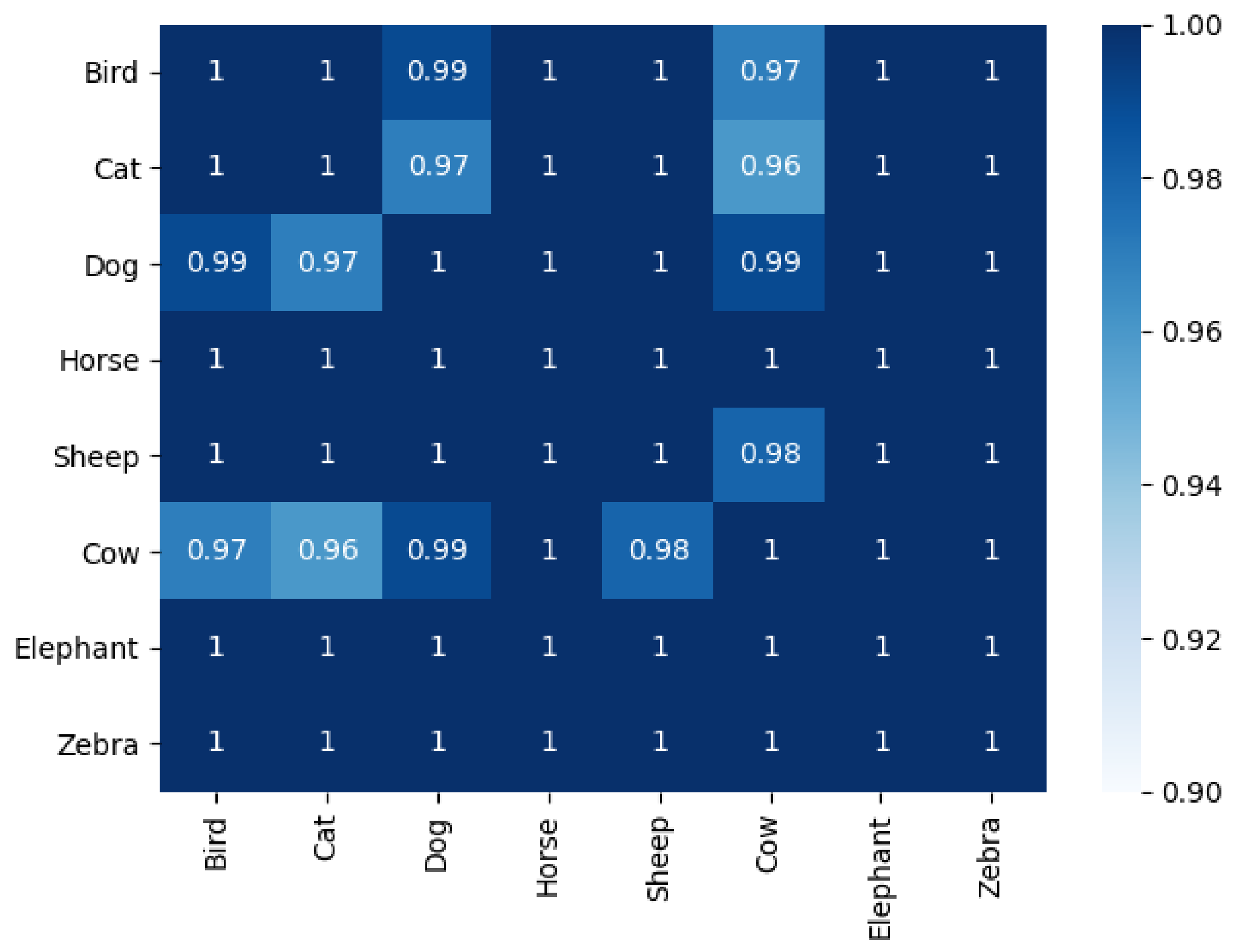} (b) TikTok
		\caption{Cosine similarity for the animal category.}
	\end{minipage}
\end{figure*}

\begin{figure*}[t]
	\centering
	\begin{minipage}{.48\columnwidth} 
		\centering
		\includegraphics[width=\textwidth, height=7cm, keepaspectratio]{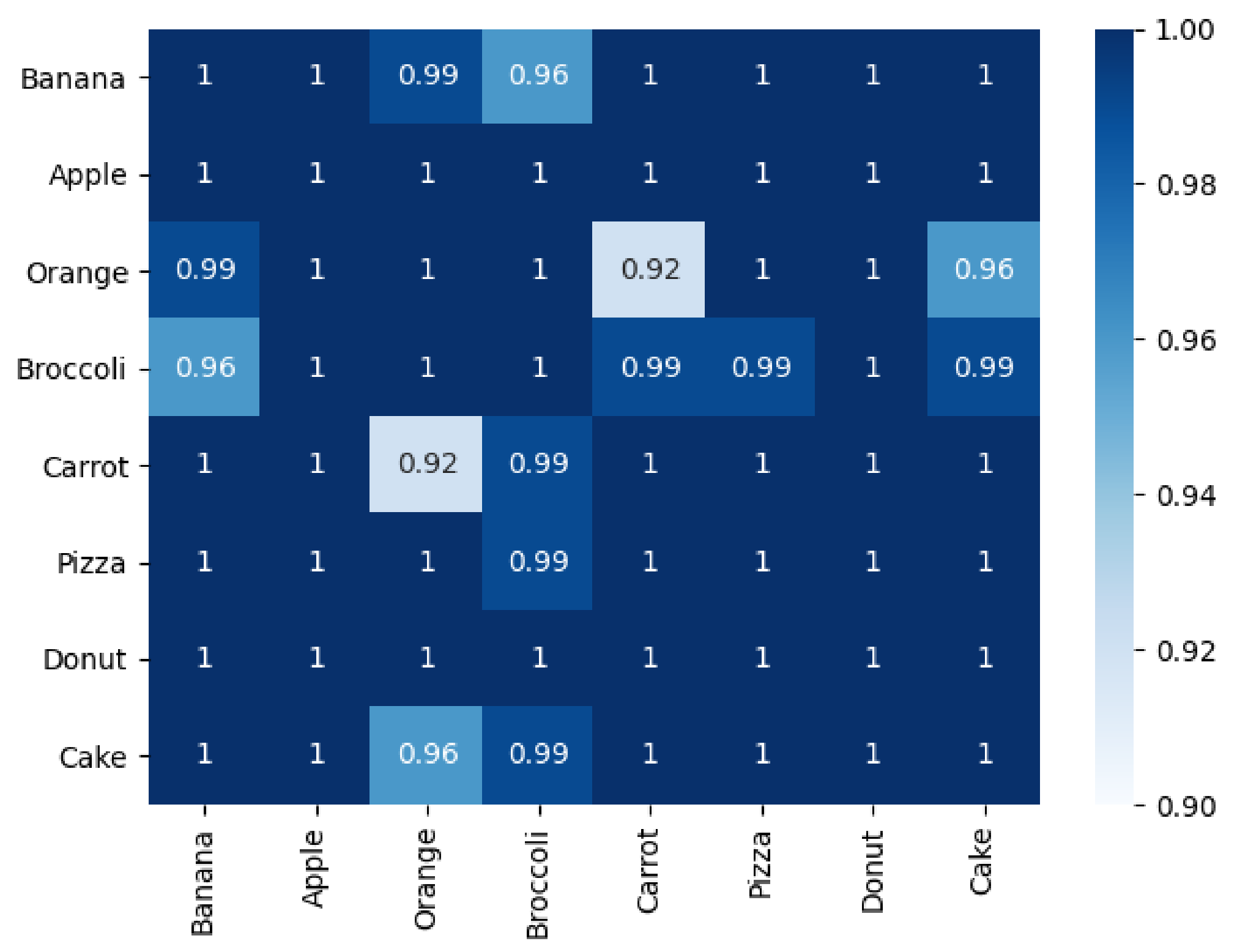} (a) Douyin
	\end{minipage}%
	\begin{minipage}{.48\columnwidth} 
		\centering
		\includegraphics[width=\textwidth, height=7cm, keepaspectratio]{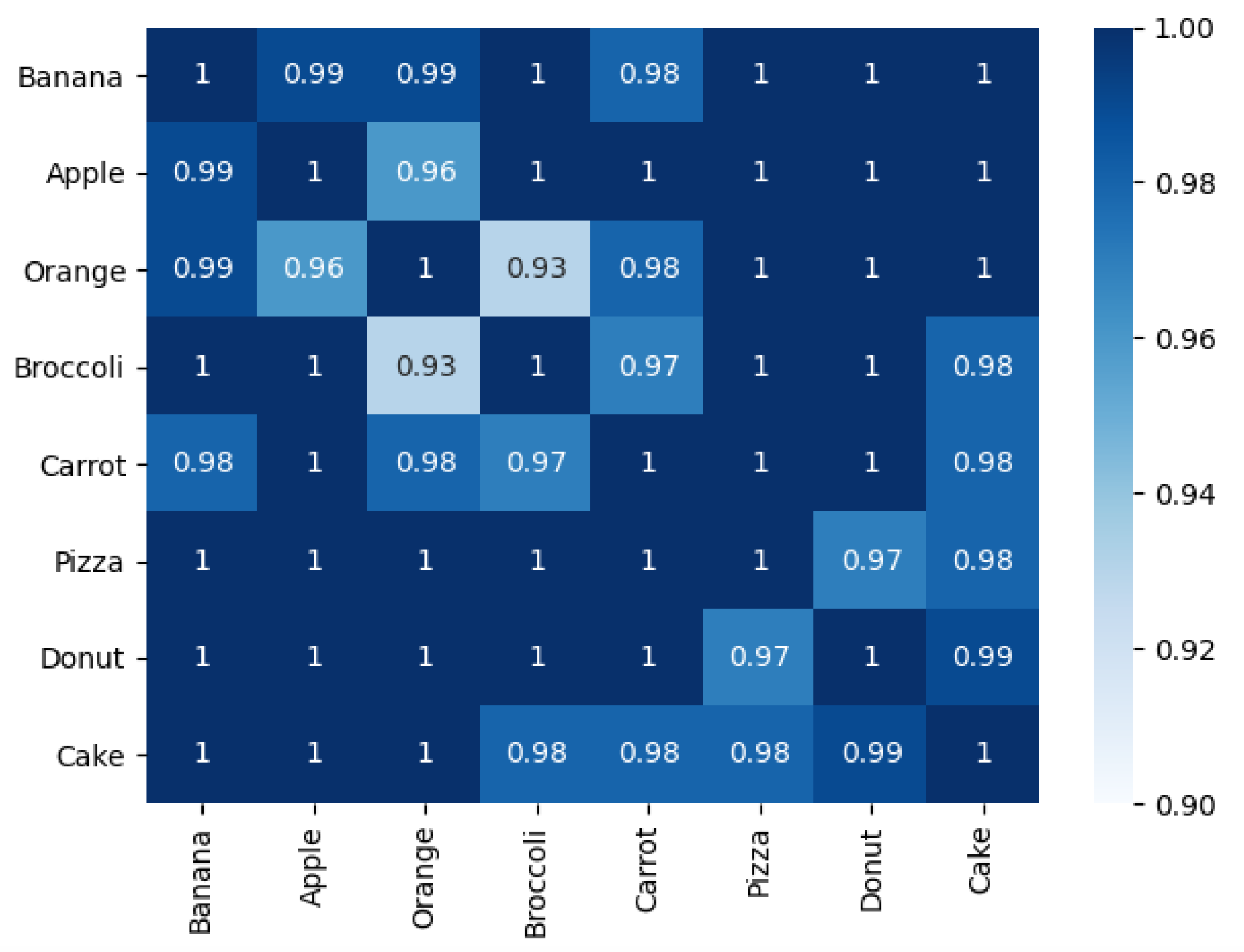} (b) TikTok
		\caption{Cosine similarity for the food category.}
	\end{minipage}
\end{figure*}

\begin{figure*}[t]
	\centering
	\begin{minipage}{.5\columnwidth} 
		\centering
		\includegraphics[width=\textwidth, height=7cm, keepaspectratio]{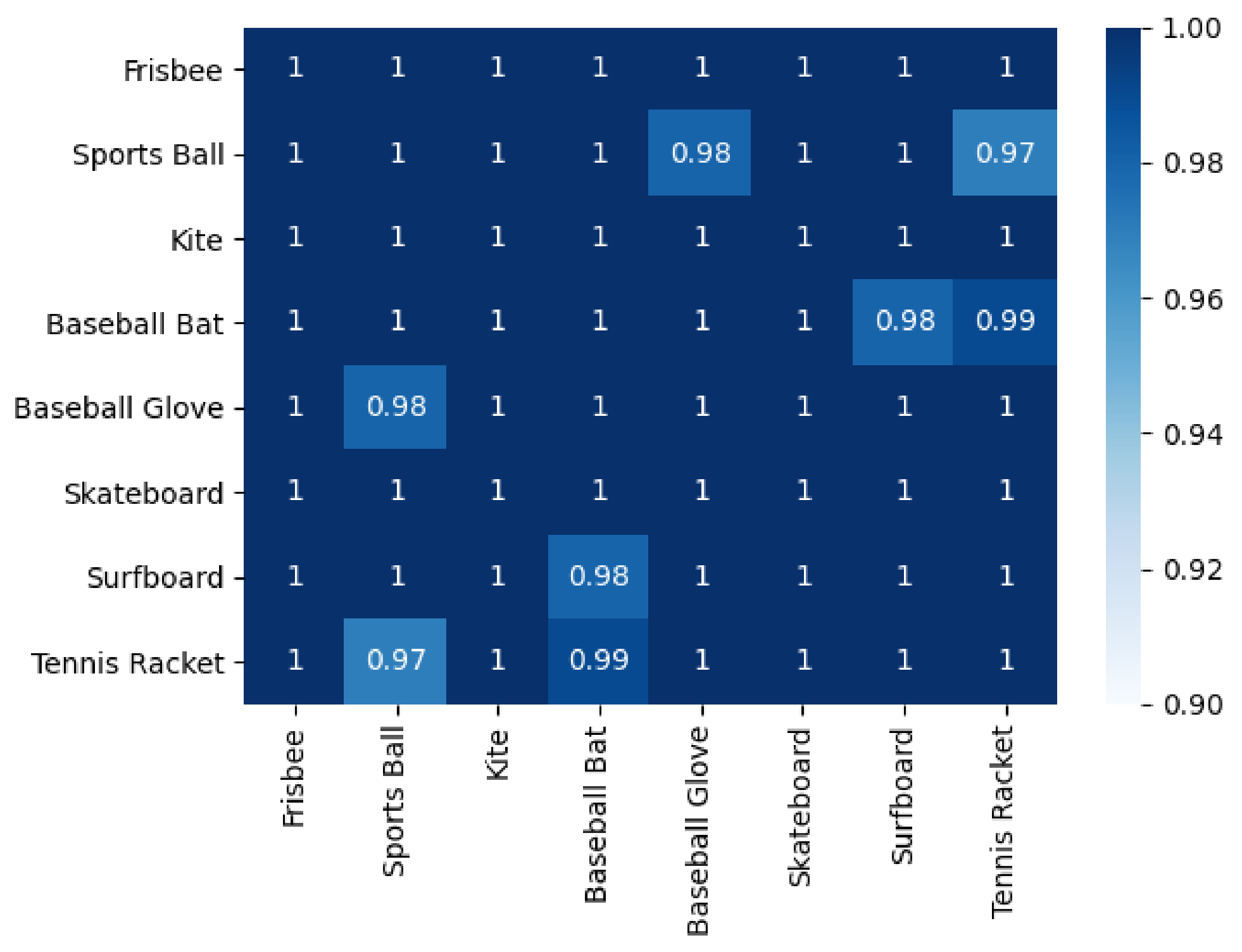} (a) Douyin
	\end{minipage}%
	\begin{minipage}{.5\columnwidth} 
		\centering
		\includegraphics[width=\textwidth, height=7cm, keepaspectratio]{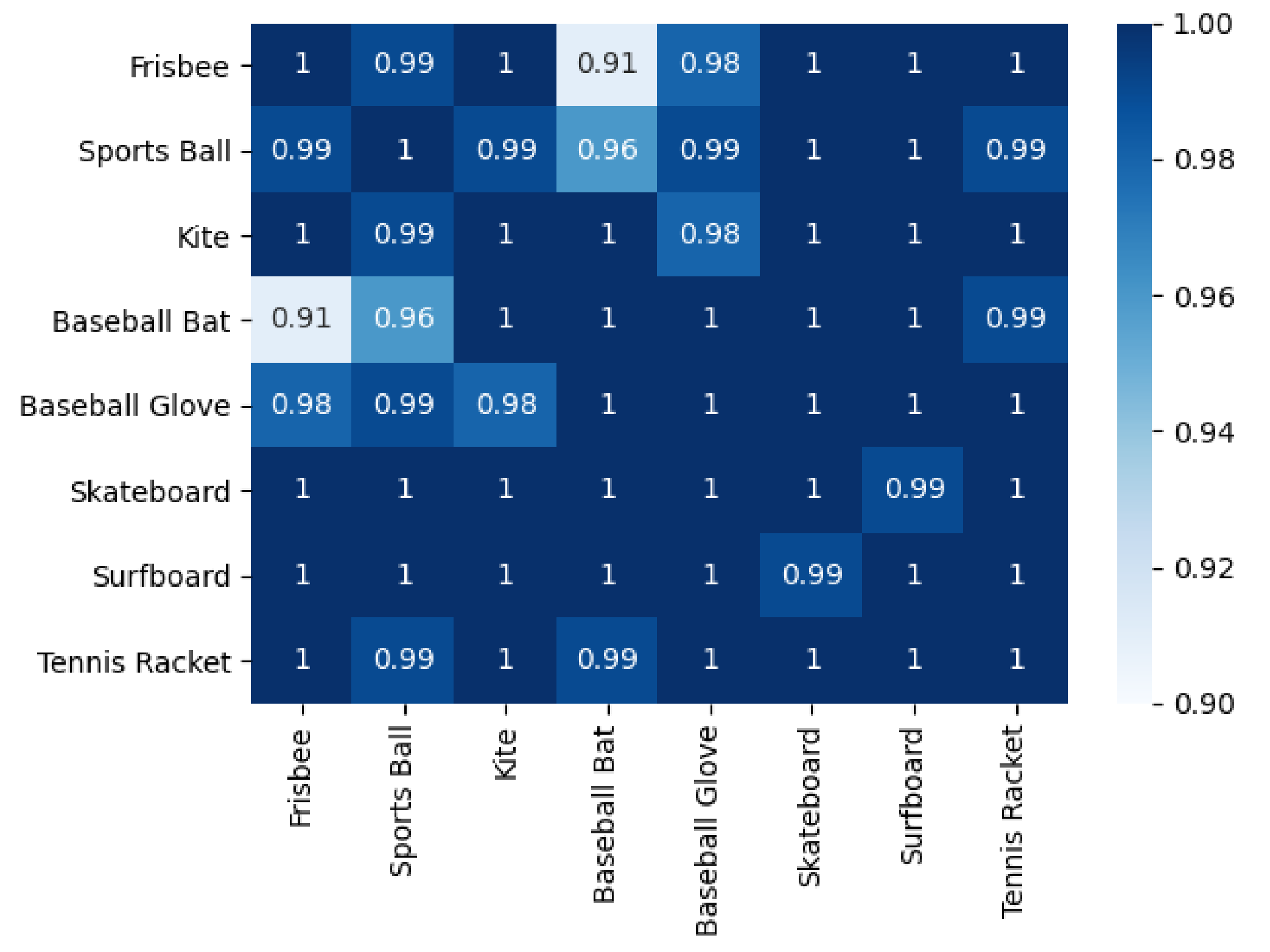} (b) TikTok
		\caption{Cosine similarity for the sports category.}
	\end{minipage}
\end{figure*}

\begin{figure*}[t]
	\centering
	\begin{minipage}{.5\columnwidth}
		\centering
		\leftskip-2.7cm
		\includegraphics[height=17cm, keepaspectratio]{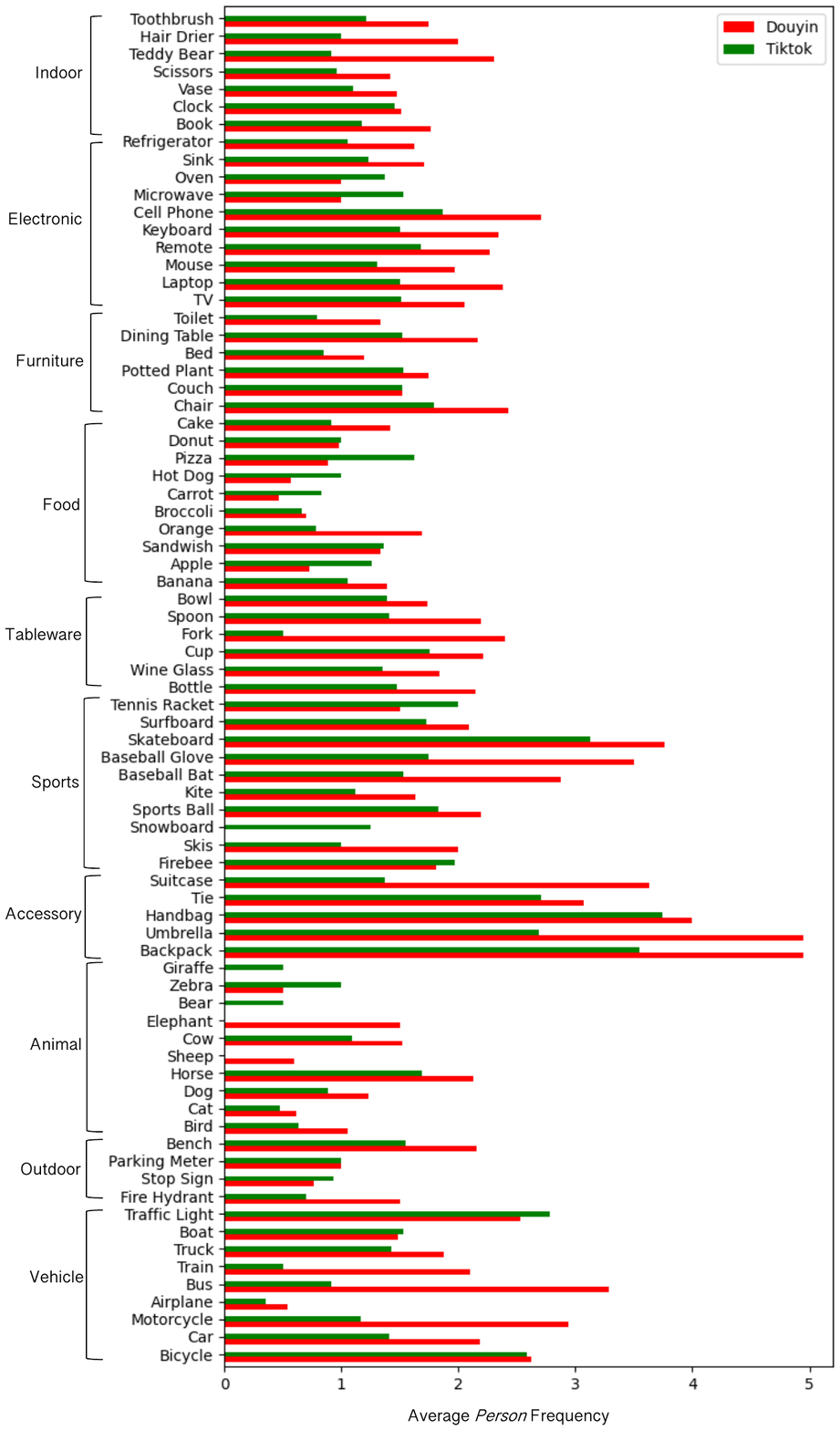}
	\end{minipage}%
	\begin{minipage}{.5\columnwidth}
		\centering
		\leftskip-2cm
		\includegraphics[height=17cm, keepaspectratio]{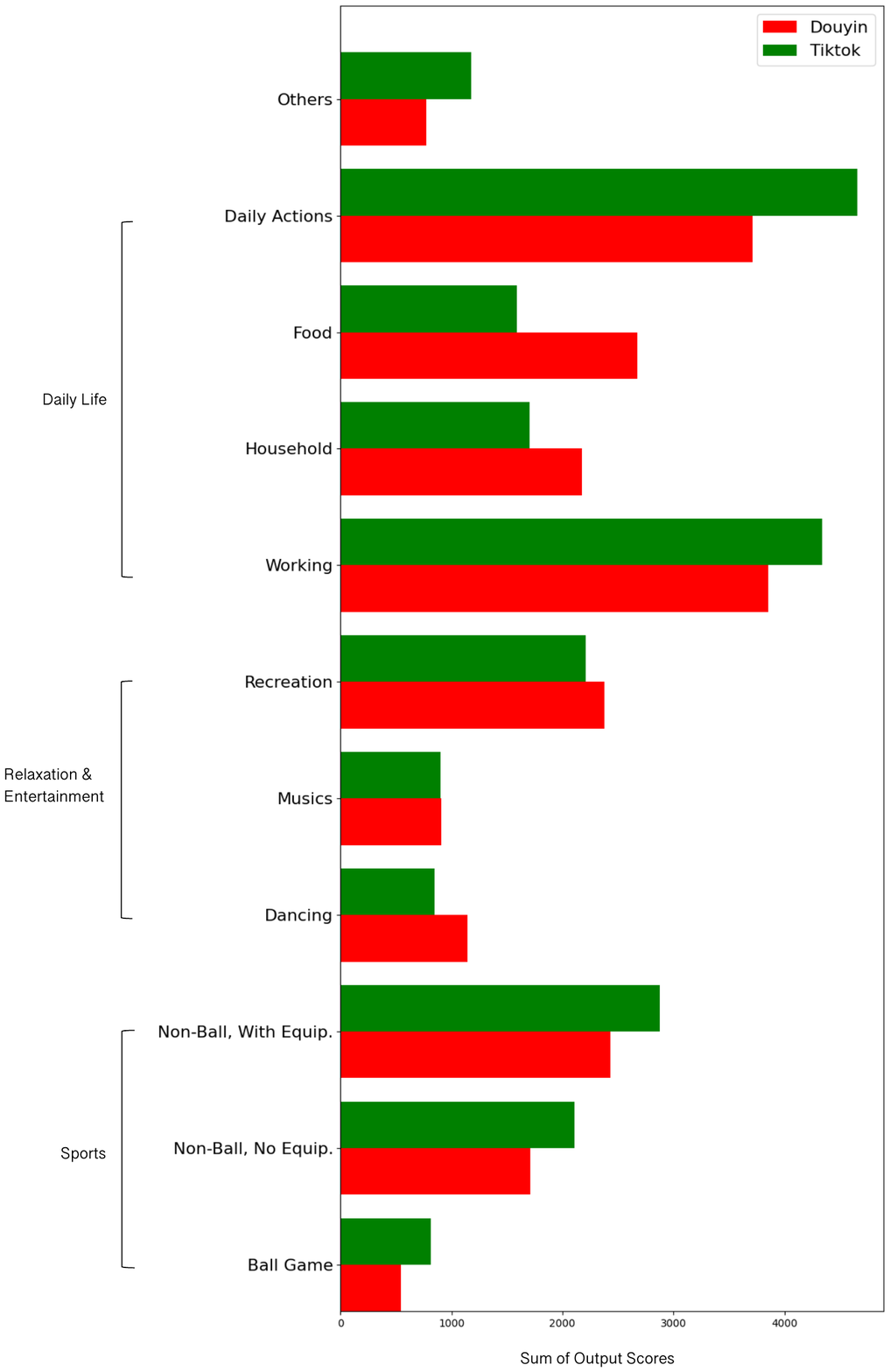}
	\end{minipage}
	\vspace{-0.4cm}
			\caption{Average Label-Person Distribution (Left) and Human Action Frequency by Category (Right).}
\end{figure*}

\section{Introduction}
Online social networks often reflect a wide variety of lifestyles through the shared contents from users. Recently, video-sharing social networks are becoming increasingly popular due to the advances in internet and video capture technologies. This new form of social media enables users to share their interests and daily lives more effectively by carefully composing short videos for maximum effect. The content of videos, for example, the main objects, the background, and the actions in turn would provide an accurate representation of the culture, fashion, and value within a specific geographic region. Therefore, videos from various locations are expected to exhibit different cultural characteristics.

Our study aims at understanding the cultural differences between TikTok and Douyin because these two applications offer similar functions but for different markets. Both applications have a 15-second default length and the restriction can be relaxed to a maximum of 5 minutes with special applications or invitations. They also concentrate on dynamic expressions through lively actions, creative challenges, and so on. The target users in Douyin are primarily from China, especially the youth between 15 to 25 years old. TikTok is the international version of Douyin, as a separate system available in the app store for people outside China. Although they are prevalent in many countries, our study only focuses on investigating the cultural differences in the United States and China.

In this study, cultural preference is investigated by conducting object label analysis and human action analysis. For object label analysis, we categorize the labels computed by Faster R-CNN \cite{b1} and analyze in terms of three aspects: label statistics, label similarity, and label-person distribution. Figure 1 shows the label frequency distribution. The set of selected labels together describes the main content of the video, and the collection of main contents indicates the characteristics of a culture. Furthermore, we apply I3D \cite{b2} to detect video actions and categorize them in a similar manner (Figure 2).

First, we summarize and compare the basic statistics for the frequency of various labels in TikTok and Douyin. Based on the quantity and variance of labels, we find that although some labels possess similar distributional patterns, differences among high-frequency labels exist as the reflection of the cultural preference in each group.

Next, we use the label similarity to measure cultural diversity by examining the difference across labels. The sum of label differences can be represented in a similarity matrix using the Cosine similarity. It enables us to investigate the diversity both within and across categories.

Furthermore, we focus on the average frequency distribution of {\itshape person} condition on each label, because {\itshape person} is the most frequent and important label. There is an intimate cause-effect relation between the count of {\itshape person} and the appearance of different objects, and the collection of those relations reflects the cultural difference. 

Finally, since social videos are often centered around human actions, automatic human action analysis is performed to observe the dynamic aspects of cultural difference. With the help of frequency statistics, we discover that users in different countries and cultures tend to have different expression habits and behave differently.

The significance of our research about TikTok and Douyin stems from the extremely high popularity of these two applications with similar media forms but strikingly different primary markets in the United State and China, respectively. Given the offering of two market-tailored versions, each market is independent and thus relatively closed with respect to the content of the video stories. Therefore, when the two applications grow in different regions, will the content reflect the local conditions? Content shifts according to users’ unique habits and values, which are shaped by different cultures and backgrounds. With such inspiration in mind, we formulate the following research questions: (1) Does the distributions of video labels reflect cultural differences? (2) How different are the distributions across labels? (3) What is the correlation between person and other labels? and (4) Do users with different cultural backgrounds behave differently?

\section{Related Work}
Many recent studies in the emerging computational social science deviate from the traditional research by directly analyzing the visual content and the associate cultural factors through automated content analysis. Furthermore, the early published works related to short video data mining primarily focused on video metadata indirectly rather than video content directly. Therefore, our approach breaks these two traditional paradigms through the parallel use of large-scale quantitative analysis and automated visual content analysis on different cultural subjects and backgrounds.

Our social media analysis is video-based, which is closely related to the work by Zhang et al. on label-based analysis using a similarity matrix and object distributions by region \cite{b3}. Rather than using only the video thumbnails, our method is facilitated by understanding the full content of the video collections using the state-of-the-art computer vision techniques described by Jha et al. \cite{b4}. Other related research also adopted similar technology, such as anorexia on YouTube \cite{b5}, background modeling \cite{b6}, lingual orthodontic treatment \cite{b7}, and video object segmentation \cite{b8}.

In a related image-based cultural study, You et al. uncovered spatiotemporal trends \cite{b9}. Image formation technology was used to measure tourism information quality on Sina Weibo \cite{b10}. Amato et al. implemented a convolutional neural network (CNN) to perform social media food image recognition on Twitter \cite{b11}. CNN is a specific type of Deep Neural Networks (DNN) in deep learning \cite{b18}. Another research by Pittman and Reich is oriented toward studying loneliness, where they compared image-based and text-based social media platforms \cite{b12}. Lai et al. also used image data to predict the personality traits, needs, and values of social media users \cite{b13}.

The existing research about Douyin was mainly concerned with users’ motivation, expectation, and behavior, including user interface analysis \cite{b14} and online network sampling \cite{b15}, which combined TikTok and Douyin as a whole. In contrast, our work aims to investigate the unique territory of cross-culture comparison as far as social media video is concerned, taking advantage of the natural separation created by having two different versions of essentially the same social video sharing application. 

\section{Data}
We collect video data by randomly downloading 5,000 videos for TikTok and 5,000 videos for Douyin. To ensure that the sampling process is sufficiently random, videos are downloaded directly from the “trending” section without signing into the application (thus free from the filtering and recommendation for each user). Different videos have different lengths, and thus are composed of different numbers of frames. Each second contains 30 frames. For each video, we capture one frame every five seconds. In total, there are 16,063 frames for TikTok and 15,217 frames for Douyin. Since the target market of Douyin is China, most downloaded Douyin videos come from China. In contrast, TikTok targets the entire world. In this study, we only focus on the videos from the United States by restricting the IP address.



\section{Methodology}
In contrast to prior work, we are interested in understanding the content difference from captured video frames. To evaluate the objects in frames, we first implement Faster R-CNN \cite{b1} and pre-train it with the MS-COCO dataset. MS-COCO collects labeled common objects in daily scenes to promote precise object localization \cite{b16}. Faster R-CNN predicts a list of predefined objects and provides a confidence score for each object. The higher the score is, the more likely the prediction is correct. We set the accepting threshold to 0.85 (every object below this threshold is ignored) to strike a balance between information accuracy and data richness. In other words, every object above 0.85 will be counted as 1 and otherwise as 0. Next, to normalize the data, we divide the total object frequency by the number of frames.

There are 92 labels \cite{b16} in the MS-COCO dataset, as shown in Table 1. We filter out labels that are not detected for both Douyin and TikTok and retain 80 labels. We empirically group the labels into 11 categories \cite{b16}.  
First, we compare the label frequency both within and across categories. Since different labels exhibit high frequency differences, we take the log to normalize the data. Label mean and range are also compared by the categories in Table 1. Second, we create similarity matrices calculated by the Cosine similarity to illustrate the label differences. Finally, for each label, we calculate the average {\itshape person} frequency and analyze the distribution. 

For videos that contain {\itshape person}, we also implement the I3D algorithm \cite{b2} to analyze human actions. The algorithm outputs a score for each action to represent the occurrence probability for every 0.64 seconds. We use the average probability of all video frames as the video prediction. If an action has a probability larger than 0.04 (empirically determined), we regard the video as containing such an action. There are 400 actions (Figure 1) in the Kinetics dataset \cite{b17}. We use the hierarchical taxonomy to separate actions into 11 categories and 3 super-categories. We then analyze the frequency distribution of these categories and super-categories for TikTok and Douyin.

\section{Analysis Results}

\subsection{Object Label Analysis}

\subsubsection{Label Statistics}

In general, we analyze the label characteristics by the normalized frequency shown on the left of Figure 4. We find some patterns in the categories. First, for all outdoor labels, TikTok dominates Douyin. It indicates TikTok users like to shoot videos outdoors. For most labels in tableware, Douyin leads in their frequencies. However, TikTok leads in electronics. This shows that a Douyin user's life is more mundane, in contrast to TikTok which is more technological. There is a mix in animal, with more livestock in Douyin but more pets in TikTok. Such animal related differences could represent unique cultural characteristics in a specific region. 

Next, we examine the average frequency of labels (the appearance frequency in each video divided by the number of captured frames) as shown on the right of Figure 4 which shows more fine-grained comparative statistics. We discover that the mean is relatively stable, but the range varies dramatically. This implies that label distributions are skewed with outliers on the high-frequency side. Most means are smaller than 1.0 since they are computed for each label in each video and every frame we take includes that label. We first analyze {\itshape bicycle} and {\itshape boat} that may represent a certain lifestyle. {\itshape Bicycle} in TikTok has a range far greater than that in Douyin but for {\itshape boat} this is slightly reversed. It implies that they have different vehicle preferences during commuting or exercise.

Other interesting labels are worth further discussion. For example, {\itshape cell phone} and {\itshape book} show higher variance in TikTok than Douyin, despite little difference in the mean. Perhaps people in Douyin use them for multiple purposes, such as game watching, story telling, or exhibition. 

\subsubsection{Label Similarity}

We analyze the relationship between labels and select several representative categories. Such relationships are captured by the similarity matrices with the Cosine Similarity shown in Figures 6 to 8. Note that we focus more on the general pattern than individual numbers, since simultaneous appearance of two labels in one category is rare, but repeated rare events in a category is often not coincidence. For easy comparison, the color scale remains the same for every matrix, which is between 0.9 and 1.

Figure 5 correlates animals in Douyin and TikTok. We can see Douyin shows more numbers smaller than 1. It means animals in Douyin are more likely to appear together, especially for {\itshape sheep} and {\itshape bird}. This grouping behavior happens with big farms, popular animal parks, or even village streets. Douyin users treat animals as similar subjects while TikTok users tend to grant them with individual characteristics.

Food in Figure 6 reveals an opposite distribution. TikTok users prefer to combine different foods and take videos together (e.g. {\itshape Broccoli} and {\itshape Orange}). However, Douyin users have different food combinations. A pattern mismatch between numbers smaller than 1 shows a difference in convention between Douyin and TikTok. Users in different countries have different tastes and food preferences.

This pattern remains for sports in Figure 7. More small numbers in TikTok means frequent and mixture of sports activity in groups. For example, they can use {\itshape baseball bat} to catch {\itshape frisbee} or {\itshape sports ball}. This flexible combination shows TikTok users tend to enjoy a complex and sports-heavy life.

\subsubsection{Label-Person Distribution}

In this section, we examine each label's average {\itshape person} frequency (Left of Figure 8). Average {\itshape person} frequency for a label is the total {\itshape person} frequency in frames with the label, divided by number of frames with the label. Overall, we discover that Douyin videos dominate in {\itshape person} frequency, with more frames that have more than two people. Tableware alone expresses stronger dominance (e.g. {\itshape cup} and {\itshape knife}). Their collective use may indicate that family members or friends meet together to have meals or union activities.

Some labels in other categories are also interesting, like {\itshape backpack} and {\itshape handbag} in accessory. They outnumber other labels in TikTok, but are infrequent in Douyin. The fact their frequency are higher than 1 may be due to  street snapshots and outdoor group activities. Douyin users often capture larger groups of people or squares with high density of people.

There is no obvious dominance in food. {\itshape Cake} leads in Douyin while {\itshape apple} leads in TikTok, with {\itshape donut} showing little difference. There is a mix in culture based on the user's taste and preference. Food origin and their local custom decide a specific food popularity.

\subsection{Human Action Analysis}

Since {\itshape person} is the most frequent label, we analyze in detail their activities in videos. We discover that users like {\itshape dancing} and {\itshape driving car} in the video for both platforms. However, some general frequent actions are different. For example, people in the video like to have {\itshape barbeque} and {\itshape dining} in Douyin but like to do {\itshape spray painting} and {\itshape writing} in TikTok.
We also divide actions into categories and compared their frequency differences sorted by category and super-category on the right of Figure 8. In general, working and daily actions are the two most frequent categories. Among these categories, Douyin has a clear lead in food and household, while TikTok leads in working and daily actions. The former two usually happen in a family and the latter two are more of a self-reflection. This implies that users in Douyin value families while users in TikTok value an independent lifestyle.

At a higher level of the taxonomy, TikTok users are more likely to shoot sports videos, but people in Douyin are more likely to show relaxation and entertainment activities. It indicates that TikTok users are more willing to share the moments when they are doing sports as a self-expression, but Douyin users are more casual and like to share their leisure moments.

\section{Conclusion and Discussion}
Previously, little research has been devoted to treat TikTok and Douyin as separate data source with unique video identity. In this study, we have compared the cultural difference through video content analysis of these two popular short video apps. As a social media platform that requires a relatively simple production technique, they accommodate users with various social classes and cultural backgrounds. Our study takes this advantage and comprehensively uncover user interests, with two important general findings concluded below.

First, on average, Douyin users show a simpler and more static lifestyle than TikTok. Similarity matrix analysis suggests that the labels in TikTok for static and daily-use objects are more likely to appear together under the same category.  TikTok users show a tendency to capture diverse items, while  many videos in Douyin consist of a clear subject and content focus.

Second, family events dominate in Douyin in comparison of the individual events in TikTok. We have shown in label statistical analysis that Douyin and TikTok have dominance in different categories. Douyin has more frequent items in accessory and kitchen, while TikTok shows more in the category of electronic and appliance. The former about necessity and utensil is about the basics of life and the latter goes beyond the basics of life. The label-person distribution also confirms this finding. Videos in Douyin have more indoor items related to family members while TikTok users take videos by going outdoors to interact with friends and strangers. The hypothesis is further verified via the analysis of human actions. Sports is a major component in TikTok relative to Douyin. Relaxing activities, in contrast, appears more in Douyin than TikTok. 

\section{Limitation and Significance}
Our study detects the frequency differences of common objects between TikTok and Douyin. Such differences are due to cultural and user preferences by regions. To serve such market preferences,  media companies in the short video space may design various application features in parallel to satisfy different user needs and provide users with a better experience.

This study has certain limitations. Faster R-CNN and I3D are two common algorithms employed by previous object detection studies. They have been extensively verified and are relatively accurate. In terms of video input, models trained on the MS-COCO and Kinetics dataset provide sufficient indication of common user activities and interacted objects. Therefore, further increasing the accuracy of the classification models and including more fine-grained labels can strengthen our cultural analysis. In addition, culture may evolve over time and a longitudinal analysis may provide further insight.

A video consists of not only frames but also audio and sometimes transcripts. Video background music and audio can provide additional information for cultural understanding as they also reflects user preferences and lifestyles. Some videos also contain textual content, such as logos, brands, and chats, which can be analyzed through OCR ad natural language processing to provide additional information.

\section*{Acknowledgment}
\vspace{-0.1cm}
This research is supported in part by the New York State CoE Goergen Institute for Data Science.

\end{document}